\def\year{2022}\relax
\documentclass[letterpaper]{article} 
\usepackage{aaai22}  
\usepackage{times}  
\usepackage{helvet}  
\usepackage{courier}  
\usepackage[hyphens]{url}  
\usepackage{graphicx} 
\usepackage{natbib}  
\usepackage{caption} 
\DeclareCaptionStyle{ruled}{labelfont=normalfont,labelsep=colon,strut=off} 
\usepackage[T1]{fontenc}
\usepackage{graphicx}
\usepackage{cite}
\usepackage{multirow}
\usepackage{amsmath}
\usepackage{booktabs,subcaption,amsfonts,dcolumn}
\newcolumntype{d}[1]{D..{#1}}
\usepackage{xcolor,colortbl}
\usepackage[misc,geometry]{ifsym} 
\definecolor{Gray}{gray}{0.85}
\usepackage{pgfplotstable}
\usepackage{algorithm}
\usepackage{algorithmic}

\usepackage{newfloat}
\usepackage{listings}
\floatstyle{ruled}
\newfloat{listing}{tb}{lst}{}
\floatname{listing}{Listing}
\nocopyright
\title{Medical SAM Adapter: Adapting Segment Anything Model for Medical Image Segmentation}

\author{
Junde Wu\textsuperscript{\rm 1,2,7},
Wei Ji\textsuperscript{\rm 3},
Yuanpei Liu\textsuperscript{\rm 8},
Huazhu Fu\textsuperscript{\rm 4},
Min Xu\textsuperscript{\rm 5,7},
Yanwu Xu\textsuperscript{\rm 6},
Yueming Jin\textsuperscript{\rm 2},
}
\affiliations {
\textsuperscript{\rm 1}University of Oxford,
\textsuperscript{\rm 2}National University of Singapore,
\textsuperscript{\rm 3}University of Alberta,
\textsuperscript{\rm 4}A*STAR,
\textsuperscript{\rm 5}Carnegie Mellon University,
\textsuperscript{\rm 6}Singapore Eye Research Institute,
\textsuperscript{\rm 7}Mohamed bin Zayed University of Artificial Intelligence,
\textsuperscript{\rm 8}University of Hongkong
}

\usepackage{bibentry}

\begin{document}

\maketitle

\begin{abstract}
The Segment Anything Model (SAM) has recently gained popularity in the field of image segmentation due to its impressive capabilities in various segmentation tasks and its prompt-based interface. However, recent studies and individual experiments have shown that SAM underperforms in medical image segmentation, since the lack of the medical specific knowledge. This raises the question of how to enhance SAM's segmentation capability for medical images. In this paper, instead of fine-tuning the SAM model, we propose the Medical SAM Adapter (Med-SA), which incorporates domain-specific medical knowledge into the segmentation model using a light yet effective adaptation technique. In Med-SA, we propose Space-Depth Transpose (SD-Trans) to adapt 2D SAM to 3D medical images and Hyper-Prompting Adapter (HyP-Adpt) to achieve prompt-conditioned adaptation. We conduct comprehensive evaluation experiments on 17 medical image segmentation tasks across various image modalities. Med-SA outperforms several state-of-the-art (SOTA) medical image segmentation methods, while updating only 2\% of the parameters. Our code is released at \url{https://github.com/KidsWithTokens/Medical-SAM-Adapter}.
\end{abstract}
\section{Introduction}

Very recently, the Segmentation Anything Model (SAM) \cite{kirillov2023segment} has gained significant attention as a powerful and versatile vision segmentation model. It can generate diverse and detailed segmentation masks based on user prompts. Despite its strong performance over natural images, many recent studies also show \cite{deng2023segment,roy2023sam,he2023accuracy} that it reaches subpar performance on medical image segmentation. Making medical image segmentation interactive, such as employing techniques like SAM, holds immense clinical value. An interactive system can prioritize areas of interest as indicated by the clinicians, providing them with a more immersive and personalized experience. For instance, in a single fundus image, there are often overlapping and intricately intertwined structures such as vessels, optic disc, optic cup, and macula. Interactive segmentation can greatly assist clinicians in efficiently distinguishing target tissues from these complex structures. Considering the difficulty in acquiring large-scale annotated datasets, it becomes crucial to adopt a foundational interactive model like SAM for clinical utilization.

SAM's limited performance on medical images is due to its lack of medical-specific knowledge, including challenges like low image contrast, ambiguous tissue boundaries, and tiny lesion regions. 
The state-of-the-art (SOTA) approach to address this issue is fully fine-tuning the vanilla SAM model specifically on medical data\cite{MedSAM}, which is quite costly in terms of both computation and memory footprint. Additionally, it is doubtful whether full fine-tuning is necessary, as previous studies have shown pre-trained visual models have strong transferability to medical images \cite{raghu2019transfusion, xie2018pre}. 

In this paper, we attempt to adapt the well-trained SAM to the medical image segmentation with minimum effort. Technically, we choose to fine-tune the pre-trained SAM using a parameter-efficient fine-tuning (PEFT) technique called Adaption \cite{hu2021lora}. Adaption has been a popular and widely-used technology in natural language processing (NLP) to fine-tune the fundamental pre-trained model for various downstream tasks. The main idea of Adaption is to insert Adapter modules with partial parameters into the original model and only update a small number of additional Adapter parameters while keeping the large pre-trained model frozen.

However, directly applying the Adaption technique to the medical scenario is not that straightforward. The first challenge arises from the image modality. Unlike natural images, many medical images are 3D, such as CT and MRI scans. It is unclear how to adapt the 2D SAM model for 3D medical image segmentation. Secondly, while Adaption has been successful in NLP, there is limited research on applying it to visual models, especially interactive visual models like SAM. In interactive visual models, user-provided visual prompts play a crucial role in the final prediction. How to incorporate Adaption with these important visual prompts remains unexplored. 

To overcome these challenges, we propose a novel adaptation framework called Medical SAM Adapter (Med-SA). In Med-SA, we introduce the Space-Depth Transpose (SD-Trans) technique to achieve 2D to 3D adaptation. In SD-Trans, we transpose the spatial dimension of input embedding to the depth dimension, allowing the same self-attention blocks can process different dimensional information given different inputs. Then we propose Hyper-Prompting Adapter (HyP-Adpt) to enable prompt-conditioned adaptation, in which we use the visual prompt to generate a series of weights that can be applied to the adaptation embedding efficiently, facilitating wide and deep prompt-adaptation interactions.

We conduct comprehensive evaluation experiments cover 17 medical image segmentation tasks across various image modalities, including CT, MRI, ultrasound images, fundus images, and dermoscopic images. The results demonstrate that Med-SA outperforms both SAM and fully fine-tuned SAM (MedSAM)\cite{MedSAM} with a significant performance gap. Med-SA also surpasses several SOTA methods that are tailor-designed for medical image segmentation, such as nnUNet, TransUNet, UNetr, and Swin-UNetr. More importantly, Med-SA achieves this superior performance by updating only 2\% extra parameters of the total SAM parameters.

\begin{itemize}
\item We present the Adaption approach for general medical image segmentation. Our framework, Med-SA, is a simple yet powerful extension of the SAM architecture, substantially enhancing its capabilities for medical applications while updating a mere 2\% of the total parameters.

\item  We propose SD-Trans to enable the segmentation of high-dimensional (3D) medical data, addressing the challenge posed by medical image modalities. 

\item We propose HyP-Adpt to facilitate prompt-conditioned adaption, acknowledging the importance of user-provided prompts in the medical domain.

\item Our extensive experiments on 17 medical image segmentation tasks with various image modalities, clearly establish Med-SA's superiority over SAM and previous state-of-the-art methods. On the widely-used abdominal multi-organ segmentation BTCV benchmark, Med-SA outperforms Swin-UNetr by 2.9\%, vanilla SAM by 34.8\%, and fully-finetuned SAM (MedSAM) by 9.4\%.

\end{itemize}

\section{Related Work}
\subsection{Interactive Segmentation}
Interactive segmentation has a rich history, initially regarded as an optimization technique by researchers \cite{grady2006random, gulshan2010geodesic, kim2010nonparametric, rother2004grabcut}. The pioneering work of DIOS \cite{xu2016deep} revolutionized interactive segmentation by integrating deep learning and incorporating positive and negative clicks as distance maps. Subsequent studies \cite{li2018interactive, liew2019multiseg} focused on addressing uncertainty by predicting multiple potential results and enabling either a selection network or the user to choose among them. CDNet \cite{chen2021conditional} further enhanced interactive segmentation by incorporating self-attention to generate more consistent predictions. RITM \cite{sofiiuk2022reviving} and AccuracyNet \cite{forte2020getting} introduced the use of previous masks as inputs to enhance the robustness and accuracy of predictions. Recently, SAM \cite{roy2023sam} demonstrated the significant impact of interactive segmentation on zero-shot segmentation and emphasized its potential importance in visual foundation models. However, limited attention has been given to interactive medical image segmentation, despite its critical role in clinical practice. For instance, a single fundus image may require the segmentation of multiple targets, such as vessels, optic disc, optic cup, and macula, depending on different requirements and use cases. Our Med-SA provides an excellent starting point for interactive medical image segmentation and aims to inspire future research in this field.

\subsection{Parameter-Efficient Fine-Turning}
PEFT has proven to be an efficient strategy for fine-tuning a large, fundamental model for a specific usage \cite{zaken2021bitfit}. Compared to full fine-tuning, it keeps most of the parameters frozen and learns significantly fewer parameters, often less than 5\% of the total. This enables efficient learning with faster updates. Studies have also shown that PEFT approaches work better than full fine-tuning as they avoid catastrophic forgetting and generalize better to out-of-domain scenarios, especially in low-data regimes \cite{zaken2021bitfit}. Among all PEFT strategies, Adaption\cite{hu2021lora} stands out as an effective tool for fine-tuning large fundamental vision models for downstream tasks, not only in NLP but also in computer vision. Recent studies have shown that Adaption can be easily adopted in various downstream computer vision tasks\cite{he2022parameter, chen2022adaptformer}. Therefore, we believe Adaption is the most fitting technique for carrying SAM to the medical domain. We anticipate that this simple, clean yet powerful Med-SA, will unlock greater possibilities for the development of foundational medical models. 





\section{Method}
\subsection{Preliminary: SAM architecture}
To begin with, we provide an overview of the SAM architecture. SAM comprises three main components: an image encoder, a prompt encoder, and a mask decoder. The image encoder is based on a standard Vision Transformer (ViT) pre-trained by MAE. Specifically, we use the ViT-H/16 variant, which employs 14×14 windowed attention and four equally-spaced global attention blocks, as shown in \ref{fig:MSA} (a). The output of the image encoder is a 16× downsampled embedding of the input image. The prompt encoder can be either sparse (points, boxes) or dense (masks). In this paper, we focus only on the sparse encoder, which represents points and boxes as positional encodings summed with learned embeddings for each prompt type. The mask decoder is a Transformer decoder block modified to include a dynamic mask prediction head. The decoder uses two-way cross-attention to learn the interaction between the prompt and image embeddings. After that, SAM upsamples the image embedding, and an MLP maps the output token to a dynamic linear classifier, which predicts the target mask of the given image.

\subsection{Med-SA architecture}\label{AA}

\begin{figure*}
    \centering
    \includegraphics[width=\linewidth]{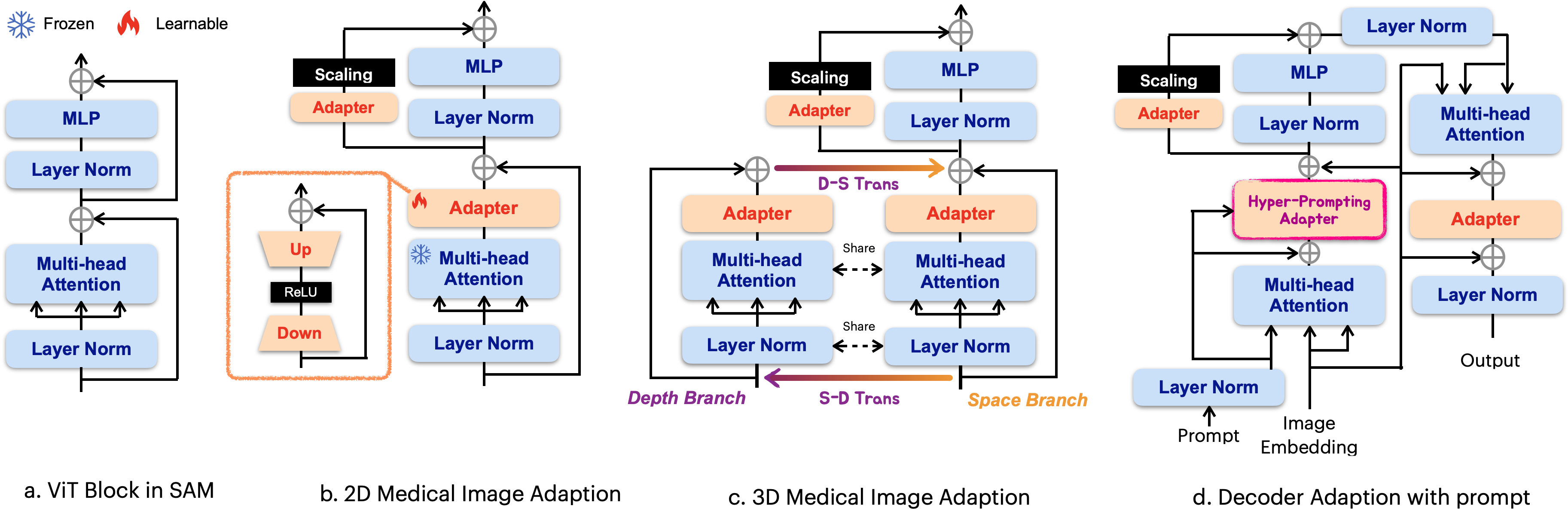}
    \caption{Med-SA architecture. We use (b) as the encoder with standard Adapter to process 2D medical images, and (c) incorporating SD-Trans to process 3D images. Then we use (d) as the decoder with HyP-Adpt to incorporate the prompts.}
    \label{fig:MSA}
\end{figure*}

\begin{figure}
    \centering
    \includegraphics[width=\linewidth]{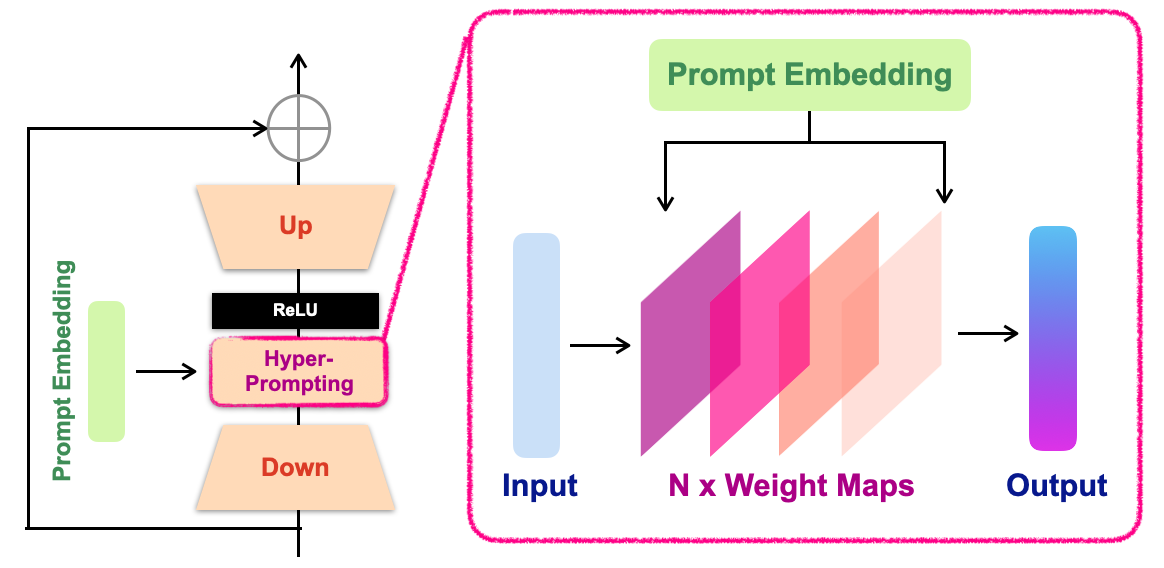}
    \caption{HyP-Adpt architecture. We utilize Prompt Embedding to generate the weights that are applied to the Adapter Embedding.}
    \label{fig:hyper}
\end{figure}

Our objective is to enhance the medical capability of the SAM architecture for medical image segmentation tasks through fine-tuning. Rather than fully adjusting all parameters, we maintain the pre-trained SAM parameters frozen, devise an Adapter module and integrate it to designated positions. The Adapter serves as a bottleneck model, consisting of a down-projection, ReLU activation, and up-projection sequentially, as illustrated in \ref{fig:MSA} (b). The down-projection compresses the given embedding into a lower dimension using a simple MLP layer, while the up-projection expands the compressed embedding back to its original dimension using another MLP layer.

In the SAM encoder, we utilize two adapters for each ViT block. For a standard ViT block (depicted in \ref{fig:MSA}(a)), the first Adapter is positioned after the multi-head attention and before the residual connection (as depicted in \ref{fig:MSA} (b)). The second Adapter is placed in the residual path of the MLP layer following the multi-head attention. Immediately after the second Adapter, we have scaled the embedding with a scale factor $s$ following \cite{chen2022adaptformer}.

In the SAM decoder, we incorporate three adapters for each ViT block. The first Adapter is employed to integrate the prompt embedding, and to achieve this, we introduce a novel structure called the Hyper-Prompting Adapter (HyP-Adpt), which is further elaborated in \ref{sec:hyper}. 
The second Adapter in the decoder is deployed in exactly the same way as in the encoder, to adapt the MLP-enhanced embedding. The third Adapter is deployed after the residual connection of the image embedding-to-prompt cross-attention. Another residual connection and layer normalization are connected after the adaption to output the final results.

\subsection{SD-Trans}
Adapting SAM to medical image segmentation poses a challenge due to the dimensional disparity between 2D images and the prevalent 3D modalities like MRI and CT scans. In clinical usage, understanding the correlation between slices is crucial for accurate decision-making. While SAM can be applied to each slice of a volume to obtain the final segmentation, it fails to consider the close volumetric correlation inherent in 3D medical image segmentation, as highlighted in previous studies \cite{hatamizadeh2022unetr, hatamizadeh2022swin, xing2023diff}. To address this limitation, we propose SD-Trans, inspired by image-to-video adaptation \cite{liu2019deep}. The specific structure is depicted in \ref{fig:MSA} (c).

As shown in the image, in each block, we bifurcate the attention operation into two branches: the space branch and the depth branch. For a given 3D sample with depth $D$, we input $D \times N \times L$ into the multi-head attention of the space branch, where $N$ represents the number of embeddings, and $L$ denotes the embedding length. Here, $D$ corresponds to the number of operations, allowing the interaction to be applied over $N \times L$, capturing and abstracting spatial correlations as embeddings. In the depth branch, we transpose the input matrix to obtain $N \times D \times L$ and subsequently feed it into the same multi-head attention. Although employing the same attention mechanism, the interaction now occurs over $D \times L$, enabling the learning and abstraction of depth correlations. Finally, we transpose the results from the depth branch back to their original shape and add them to the results of the space branch, incorporating the depth information.

\begin{figure*}[t]
    \centering
    \includegraphics[width=0.85\linewidth]{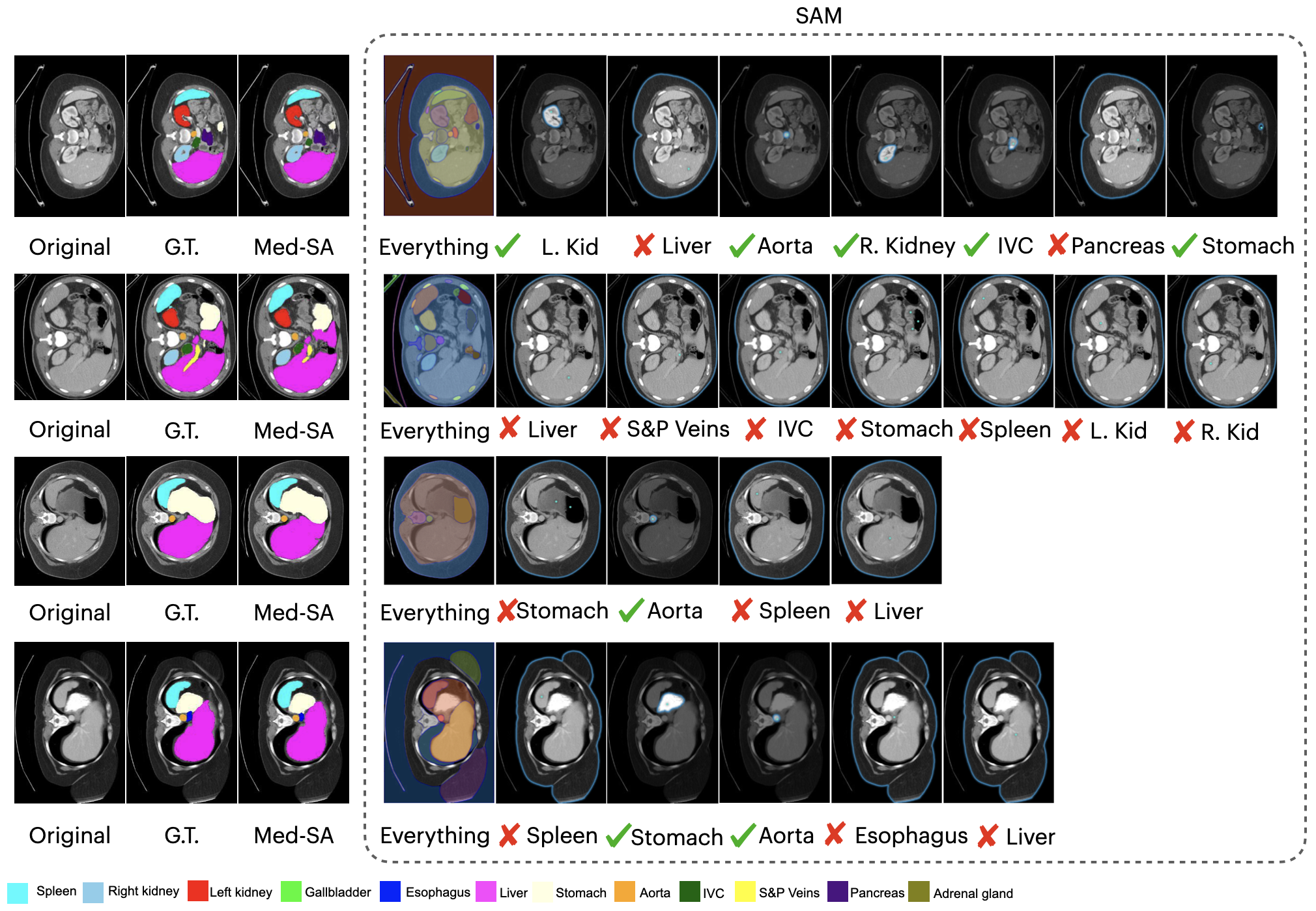}
    \caption{Visual comparison of Med-SA and SAM on abdominal multi-organ segmentation. We use Check mark to represent SAM correctly found the organ and Cross to represent it lost.}
    \label{fig:amos-vis}
\end{figure*}

\begin{table*}[h]
\centering
\caption{The comparison of Med-SA with SOTA segmentation methods over BTCV dataset evaluated by Dice Score. Best results are denoted as \textbf{bold}.}
\resizebox{\linewidth}{!}{%
\begin{tabular}{c|cc|cccccccccccc|c}
\hline
Model              & Param(M)  & \multicolumn{1}{c|}{\begin{tabular}[c]{@{}c@{}}Turnable \\ Param(M)  \end{tabular}}                             & Spleen & R.Kid & L.Kid & Gall. & Eso.  & Liver & Stom.  & Aorta & IVC  &Veins & Panc. & AG  & Avg  \\ \hline
TransUNet          & 37 & 37                                             & 0.952                        & 0.927 & 0.929 & 0.662 & 0.757 & 0.969  & 0.889 & 0.920  & 0.833 & 0.791       & 0.775     & 0.637 & 0.838 \\
EnsDiff                & 32 & 32                                         & 0.938                        & 0.931 & 0.924 & 0.772 & 0.771 & 0.967  & 0.910 & 0.869  & 0.851 & 0.802       & 0.771     & 0.745 & 0.854      \\
SegDiff              & 32 & 32                                           & 0.954                        & 0.932 & 0.926 & 0.738 & 0.763 & 0.953  & 0.927 & 0.846  & 0.833 & 0.796       & 0.782     & 0.723 & 0.847     \\
UNetr                & 104 & 104                                            & 0.968                        & 0.924 & 0.941 & 0.750 & 0.766 & 0.971  & 0.913 & 0.890  & 0.847 & 0.788       & 0.767     & 0.741 & 0.856 \\
Swin-UNetr             & 138 & 138                                         & 0.971                        & 0.936 & 0.943 & 0.794 & 0.773 & 0.975  & 0.921 & 0.892  & 0.853 & 0.812       & 0.794     & 0.765 &  0.869    \\
nnUNet                & 16 & 16                                           & 0.942                        & 0.894 & 0.910 & 0.704 & 0.723 & 0.948  & 0.824 & 0.877  & 0.782 & 0.720       & 0.680     & 0.616 & 0.802 \\
\hline
SAM 1 points & 636 & 0   & 0.518  & 0.686 & 0.791 & 0.543 & 0.584 & 0.461  & 0.562 & 0.612  & 0.402 & 0.553       & 0.511     & 0.354 & 0.548     \\
SAM 3 points & 636 & 0 & 0.622  & 0.710 & 0.812 & 0.614 & 0.605 & 0.513  & 0.673 & 0.645  & 0.483 & 0.628       & 0.564     & 0.395 & 0.631     \\ 
SAM BBox 0.75 & 636 & 0 & 0.415  & 0.621 & 0.678 & 0.580 & 0.595 & 0.469  & 0.521 & 0.612  & 0.539 & 0.655       & 0.588     & 0.327 & 0.550     \\ 
SAM BBox 0.5 & 636 & 0 & 0.346  & 0.585 & 0.592 & 0.375 & 0.426 & 0.377  & 0.451 & 0.536  & 0.392 & 0.576       & 0.426     & 0.202 & 0.440     \\ 
MedSAM 1 point & 636 & 636 & 0.751  & 0.814 & 0.885 & 0.766 & 0.721 & 0.901  & 0.855 & 0.872  & 0.746 & 0.771       & 0.760     & 0.705 & 0.803     \\ 
MedSAM 3 points & 636 & 636 & 0.758  & 0.831 & 0.889 & 0.782 & 0.733 & 0.917  & 0.858 & 0.876  & 0.755 & 0.776       & 0.763     & 0.716 & 0.820     \\ 
MedSAM BBox 0.75  & 636 & 636 & 0.746  & 0.842 & 0.873 & 0.772 & 0.745 & 0.897  & 0.860 & 0.889  & 0.743 & 0.745       & 0.739     & 0.701 & 0.804     \\
MedSAM BBox 0.5 & 636 & 636 & 0.621  & 0.736 & 0.801 & 0.721 & 0.715 & 0.811  & 0.714 & 0.770  & 0.622 & 0.618       & 0.630     & 0.545 & 0.692     \\ \hline
Med-SA 1 point   & 636 & 13                                               & 0.978                        & 0.935 & 0.966 & 0.823 & 0.818 & 0.981  & 0.931 & 0.915  & 0.877 & 0.811       & 0.767     & 0.809 &  0.883   \\ 
Med-SA 3 points   & 636 & 13                                               & 0.980                        & 0.936 & 0.968 & 0.826 & \textbf{0.821} & \textbf{0.986}  & 0.934 & 0.917  & 0.878 & 0.813       & 0.771     & 0.818 &  0.887   \\ 
Med-SA BBox 0.5   & 636 & 13                                               & 0.954                        & 0.910 & 0.952 & 0.810 & 0.807 & 0.975  & 0.928 & 0.912  & 0.868 & 0.809       & 0.769     & 0.813 &  0.876   \\ 
Med-SA BBox 0.75   & 636 & 13                                              & \textbf{0.985}                        & \textbf{0.947} & \textbf{0.975} & \textbf{0.842} & 0.808 & 0.983  & \textbf{0.942} & \textbf{0.939}  & \textbf{0.899} & \textbf{0.852}       & \textbf{0.790}     & \textbf{0.823} &  \textbf{0.898}   \\ \hline

\end{tabular}}\label{tab:btcv}
\end{table*}

\subsection{HyP-Adpt}\label{sec:hyper}
While adaptation techniques have been applied to visual models in a few previous works, the application of adaptation to interactive visual models remains largely unexplored. The interactive behavior between the source task and the downstream task can exhibit significant differences. Therefore, it becomes crucial to incorporate the visual prompt, which plays a key role in the interactive model, into the adapter. In this regard, we propose a solution called HyP-Adpt, aimed at achieving prompt-conditioned adaptation.

The idea behind HyP-Adpt is inspired by HyperNetworks \cite{ha2016hypernetworks}, which employ one network to generate weights for another network for the knowledge conditioning. We adopt the high-level concept of HyperNetworks but redesign it to efficiently apply it at the feature level. Specifically, we utilize only projection and reshaping operations to generate a sequence of weight maps from the prompt embedding. These weight maps are then directly applied (matrix product) to the adapter embedding. This approach enables wide and deep feature-level interaction while also significantly reducing the number of parameters required, as compared to generating an entire network.

Specifically, we conduct the hyper-prompting over the reduced embedding of the Adapter $e^{down}$. In the mean time, the prompt information (click location, click attribution, or bounding box location) is concatenated and reduced as prompt embedding $e^{prompt}$. Then we use $e^{prompt}$ to generate the a sequence of weights, taking one of it to illustrate, it can be represented as:
\begin{equation}
    W = Re (M (e^{prompt})),
\end{equation}
where $Re$ denotes reshape, and $M$ denotes the MLP layer to project $e^{prompt} \in \mathcal{R}^{N \times L}$ to $e^{prompt} \in \mathcal{R}^{N \times (L^{in} * L^{out})}$, in which $*$ is value multiplication, $L^{in}$ of the first weight will be the length of $e^{down}$, and $L^{out}$ of the last weight will be the target length of the output. After that, we reshape $e^{prompt}$ from 1D embedding to 2D weight $w^{prompt} \in \mathcal{R}^{N \times L^{in} \times L^{out}}$, and apply it over $e^{down}$, which can be represented as:
\begin{equation}
    e^{down}_{n+1} = ReLU (Norm (e^{down}_{n} \otimes w^{prompt} )),
\end{equation}
where $\otimes$ is the matrix product. We normalize the elements along the length dimension and apply ReLU activation after then. We set 3 layers for the hyper-prompting, each weight is projected by individual MLP layers. HyP-Adpt helps to turn the parameter conditioned on the prompt information and be more flexible to different modalities and downstream tasks.

\subsection{Training Strategy}\label{AA}

\begin{table*}[h]
\centering
\caption{The comparison of Med-SA with SAM and SOTA segmentation methods on different image modalities. The grey background denotes the methods are proposed for that/those particular tasks. Performance is omitted (-) if the algorithm fails over 70\% of the samples.}
\resizebox{0.95\linewidth}{!}{%
\begin{tabular}{c|cc|cc|cc|ccc|cc|cc}
\hline
{\color[HTML]{333333} }  & \multicolumn{2}{c|}{ }    & \multicolumn{2}{c|}{{\color[HTML]{333333} Optic-Disc}}     & \multicolumn{2}{c|}{{\color[HTML]{333333} Optic-Cup}}                                                     & \multicolumn{3}{c|}{{\color[HTML]{333333} Brain-Turmor}}                                                  & \multicolumn{2}{c|}{{\color[HTML]{333333} Thyroid Nodule}}   & \multicolumn{2}{c}{{\color[HTML]{333333} Melanoma}}                                              \\ \hline
{\color[HTML]{333333} }    & Param(M)               & \multicolumn{1}{c|}{\begin{tabular}[c]{@{}c@{}}Turnable \\ Param(M)  \end{tabular}}   & {\color[HTML]{333333} Dice}                         & {\color[HTML]{333333} IoU}     & {\color[HTML]{333333} Dice}                         & {\color[HTML]{333333} IoU}                          & {\color[HTML]{333333} Dice}                         & {\color[HTML]{333333} IoU} & {\color[HTML]{333333} HD95}                          & {\color[HTML]{333333} Dice}                         & {\color[HTML]{333333} IoU} & {\color[HTML]{333333} Dice}                         & {\color[HTML]{333333} IoU} \\ \hline
{\color[HTML]{333333} ResUNet} & 17 & 17 & \cellcolor[HTML]{EFEFEF}{\color[HTML]{333333} 92.9} & \cellcolor[HTML]{EFEFEF}{\color[HTML]{333333} 85.5}  & \cellcolor[HTML]{EFEFEF}{\color[HTML]{333333} 80.1} & \cellcolor[HTML]{EFEFEF}{\color[HTML]{333333} 72.3} & {\color[HTML]{333333} 78.4}                            & {\color[HTML]{333333} 71.3} & {\color[HTML]{333333} 18.71}                             & {\color[HTML]{333333} 78.3}                            & {\color[HTML]{333333} 70.7}      & {\color[HTML]{333333} 87.1}                            & {\color[HTML]{333333} 78.2}                       \\
{\color[HTML]{333333} BEAL}  & 25 & 25  & \cellcolor[HTML]{EFEFEF}{\color[HTML]{333333} 93.7} & \cellcolor[HTML]{EFEFEF}{\color[HTML]{333333} 86.1}  & \cellcolor[HTML]{EFEFEF}{\color[HTML]{333333} 83.5} & \cellcolor[HTML]{EFEFEF}{\color[HTML]{333333} 74.1} & {\color[HTML]{333333} 78.8}                            & {\color[HTML]{333333} 71.7} & {\color[HTML]{333333} 18.53}                            & {\color[HTML]{333333} 78.6}                            & {\color[HTML]{333333} 71.6}      & {\color[HTML]{333333} 86.6}                            & {\color[HTML]{333333} 78.0}                       \\ \hline
{\color[HTML]{333333} TransBTS} & 39 & 39 & {\color[HTML]{333333} 94.1}                            & {\color[HTML]{333333} 87.2}   & {\color[HTML]{333333} 85.4}                            & {\color[HTML]{333333} 75.7}                            & \cellcolor[HTML]{EFEFEF}{\color[HTML]{333333} 87.6} & \cellcolor[HTML]{EFEFEF}{\color[HTML]{333333} 78.44}  & \cellcolor[HTML]{EFEFEF}{\color[HTML]{333333} 12.44} & {\color[HTML]{333333} 83.8}                            & {\color[HTML]{333333} 75.5}       & {\color[HTML]{333333} 88.1}                            & {\color[HTML]{333333} 80.6}                       \\

{\color[HTML]{333333} EnsemDiff} & 32 & 32 & {\color[HTML]{333333} 94.3}                            & {\color[HTML]{333333} 87.8}  & {\color[HTML]{333333} 84.2}                            & {\color[HTML]{333333} 74.4}                            & \cellcolor[HTML]{EFEFEF}{\color[HTML]{333333} 88.7} & \cellcolor[HTML]{EFEFEF}{\color[HTML]{333333} 80.9} & \cellcolor[HTML]{EFEFEF}{\color[HTML]{333333} 10.85} & {\color[HTML]{333333} 83.9}                            & {\color[HTML]{333333} 75.3}       & {\color[HTML]{333333} 88.2}                            & {\color[HTML]{333333} 80.7}          \\ \hline

{\color[HTML]{333333} MTSeg}  & 27 & 27  & {\color[HTML]{333333} 90.3}                            & {\color[HTML]{333333} 83.6}    & {\color[HTML]{333333} 82.3}                            & {\color[HTML]{333333} 73.1}                            & {\color[HTML]{333333} 82.2}                            & {\color[HTML]{333333} 74.5} & {\color[HTML]{333333} 15.74}                            & \cellcolor[HTML]{EFEFEF}{\color[HTML]{333333} 82.3} & \cellcolor[HTML]{EFEFEF}{\color[HTML]{333333} 75.2} & {\color[HTML]{333333} 87.5} & {\color[HTML]{333333} 79.7}  \\

{\color[HTML]{333333} UltraUNet} & 19 & 19 & {\color[HTML]{333333} 91.5}                            & {\color[HTML]{333333} 82.8}  & {\color[HTML]{333333} 83.1}                            & {\color[HTML]{333333} 73.8}                            & {\color[HTML]{333333} 84.5}                            & {\color[HTML]{333333} 76.3}  & {\color[HTML]{333333} 14.03}                            & \cellcolor[HTML]{EFEFEF}{\color[HTML]{333333} 84.5} & \cellcolor[HTML]{EFEFEF}{\color[HTML]{333333} 76.2}   & {\color[HTML]{333333} 89.0}  & {\color[HTML]{333333} 81.8}  \\ \hline

 {\color[HTML]{333333} FAT-Net}  & 75 & 75  & {\color[HTML]{333333} 91.8}                            & {\color[HTML]{333333} 84.8}    & {\color[HTML]{333333} 80.9}                            & {\color[HTML]{333333} 71.5}                            & {\color[HTML]{333333} 79.2}                            & {\color[HTML]{333333} 72.8} & {\color[HTML]{333333} 17.35}                            & {\color[HTML]{333333} 80.8} & {\color[HTML]{333333} 73.4} & \cellcolor[HTML]{EFEFEF}{\color[HTML]{333333} 90.7} & \cellcolor[HTML]{EFEFEF}{\color[HTML]{333333} 83.9} \\

{\color[HTML]{333333} BAT} & 88 & 88 & {\color[HTML]{333333} 92.3}                            & {\color[HTML]{333333} 85.8}  & {\color[HTML]{333333} 82.0}                            & {\color[HTML]{333333} 73.2}                            & {\color[HTML]{333333} 79.6}                            & {\color[HTML]{333333} 73.5}  & {\color[HTML]{333333} 15.49}                            & {\color[HTML]{333333} 81.7} &{\color[HTML]{333333} 74.2}  & \cellcolor[HTML]{EFEFEF}{\color[HTML]{333333} 91.2}  & \cellcolor[HTML]{EFEFEF}{\color[HTML]{333333} 84.3}  \\ \hline

{\color[HTML]{333333} SegDiff} & 32 & 32  & \cellcolor[HTML]{EFEFEF}{\color[HTML]{333333} 92.6}                         & \cellcolor[HTML]{EFEFEF}{\color[HTML]{333333} 85.2}         & \cellcolor[HTML]{EFEFEF}{\color[HTML]{333333}82.5}                         & \cellcolor[HTML]{EFEFEF}{\color[HTML]{333333} 71.9}                         & \cellcolor[HTML]{EFEFEF}{\color[HTML]{333333} 85.7}                         & \cellcolor[HTML]{EFEFEF}{\color[HTML]{333333} 77.0}     & \cellcolor[HTML]{EFEFEF}{\color[HTML]{333333}14.31 }                  & \cellcolor[HTML]{EFEFEF}{\color[HTML]{333333} 81.9}                         & \cellcolor[HTML]{EFEFEF}{\color[HTML]{333333}74.8}  & \cellcolor[HTML]{EFEFEF}{\color[HTML]{333333} 87.3}                         & \cellcolor[HTML]{EFEFEF}{\color[HTML]{333333} 79.4}                          \\

nnUNet      & 16 & 16       & \cellcolor[HTML]{EFEFEF}{\color[HTML]{333333} 94.7}                                                & \cellcolor[HTML]{EFEFEF}{\color[HTML]{333333} 87.3}                 & \cellcolor[HTML]{EFEFEF}{\color[HTML]{333333} 84.9 }                                                & \cellcolor[HTML]{EFEFEF}{\color[HTML]{333333} 75.1}                                                & \cellcolor[HTML]{EFEFEF}{\color[HTML]{333333}88.5}                                                & \cellcolor[HTML]{EFEFEF}{\color[HTML]{333333}80.6}    & \cellcolor[HTML]{EFEFEF}{\color[HTML]{333333}11.20}                                            & \cellcolor[HTML]{EFEFEF}{\color[HTML]{333333}84.2}                                                & \cellcolor[HTML]{EFEFEF}{\color[HTML]{333333}76.2}          & \cellcolor[HTML]{EFEFEF}{\color[HTML]{333333}90.8}                                                & \cellcolor[HTML]{EFEFEF}{\color[HTML]{333333}83.6}                                        \\

TransUNet     & 96 & 96         & \cellcolor[HTML]{EFEFEF}{\color[HTML]{333333}95.0}                                                & \cellcolor[HTML]{EFEFEF}{\color[HTML]{333333}87.7}                        & \cellcolor[HTML]{EFEFEF}{\color[HTML]{333333}85.6}                                                & \cellcolor[HTML]{EFEFEF}{\color[HTML]{333333}75.9}                                                & \cellcolor[HTML]{EFEFEF}{\color[HTML]{333333}86.6}                                                & \cellcolor[HTML]{EFEFEF}{\color[HTML]{333333}79.0}       & \cellcolor[HTML]{EFEFEF}{\color[HTML]{333333}13.74}                                          & \cellcolor[HTML]{EFEFEF}{\color[HTML]{333333} 83.5}                         & \cellcolor[HTML]{EFEFEF}{\color[HTML]{333333}75.1}        & \cellcolor[HTML]{EFEFEF}{\color[HTML]{333333}89.4}       & \cellcolor[HTML]{EFEFEF}{\color[HTML]{333333}82.2}                                             \\ 

{\color[HTML]{333333} UNetr} & 104 & 104   & \cellcolor[HTML]{EFEFEF}{\color[HTML]{333333}94.9}                                                &  \cellcolor[HTML]{EFEFEF}{\color[HTML]{333333}87.5}  & \cellcolor[HTML]{EFEFEF}{\color[HTML]{333333}83.2}                         & \cellcolor[HTML]{EFEFEF}{\color[HTML]{333333} 73.3}                         & \cellcolor[HTML]{EFEFEF}{\color[HTML]{333333} 87.3}                         & \cellcolor[HTML]{EFEFEF}{\color[HTML]{333333} 80.6}  & \cellcolor[HTML]{EFEFEF}{\color[HTML]{333333}12.81}                       & \cellcolor[HTML]{EFEFEF}{\color[HTML]{333333} 81.7}                         & \cellcolor[HTML]{EFEFEF}{\color[HTML]{333333} 73.5}           & \cellcolor[HTML]{EFEFEF}{\color[HTML]{333333}89.7}                                                & \cellcolor[HTML]{EFEFEF}{\color[HTML]{333333}82.8}                 \\

{\color[HTML]{333333} Swin-UNetr}  & 138 & 138  & \cellcolor[HTML]{EFEFEF}{\color[HTML]{333333}95.3}                                                & \cellcolor[HTML]{EFEFEF}{\color[HTML]{333333}87.9}   &\cellcolor[HTML]{EFEFEF}{\color[HTML]{333333}84.3}                         & \cellcolor[HTML]{EFEFEF}{\color[HTML]{333333}74.5}                         & \cellcolor[HTML]{EFEFEF}{\color[HTML]{333333} 88.4}                         & \cellcolor[HTML]{EFEFEF}{\color[HTML]{333333} 81.8}               & \cellcolor[HTML]{EFEFEF}{\color[HTML]{333333}11.36}           & \cellcolor[HTML]{EFEFEF}{\color[HTML]{333333}83.5}                         & \cellcolor[HTML]{EFEFEF}{\color[HTML]{333333} 74.8}       & \cellcolor[HTML]{EFEFEF}{\color[HTML]{333333}90.2}                                                & \cellcolor[HTML]{EFEFEF}{\color[HTML]{333333}83.1}                    \\

\hline
{\color[HTML]{333333} SAM 1 points} & 636 & 0  & -                                                & -   & {\color[HTML]{333333} -}                         & {\color[HTML]{333333} -}                         & \cellcolor[HTML]{EFEFEF}{\color[HTML]{333333} 63.2}                         & \cellcolor[HTML]{EFEFEF}{\color[HTML]{333333}47.6}            & \cellcolor[HTML]{EFEFEF}{\color[HTML]{333333}32.53}             & {\color[HTML]{333333} -}                         & {\color[HTML]{333333} -}           & \cellcolor[HTML]{EFEFEF}{\color[HTML]{333333}81.6}                                                & \cellcolor[HTML]{EFEFEF}{\color[HTML]{333333}70.4}                \\

{\color[HTML]{333333} SAM 3 points} & 636 & 0 & -                                                & -    & {\color[HTML]{333333} -}                         & {\color[HTML]{333333} -}                         & \cellcolor[HTML]{EFEFEF}{\color[HTML]{333333} 71.3}                         & \cellcolor[HTML]{EFEFEF}{\color[HTML]{333333} 64.5}            &  \cellcolor[HTML]{EFEFEF}{\color[HTML]{333333}28.74}             & {\color[HTML]{333333} -}                         & {\color[HTML]{333333} -}          & \cellcolor[HTML]{EFEFEF}{\color[HTML]{333333} 85.8 }                                               & \cellcolor[HTML]{EFEFEF}{\color[HTML]{333333}77.5 }                \\

{\color[HTML]{333333} SAM BBox 0.5} & 636 & 0 & -                                                & -    & {\color[HTML]{333333} -}                         & {\color[HTML]{333333} -}                         & \cellcolor[HTML]{EFEFEF}{\color[HTML]{333333} 51.2}                         & \cellcolor[HTML]{EFEFEF}{\color[HTML]{333333} 44.6}            & \cellcolor[HTML]{EFEFEF}{\color[HTML]{333333}38.56 }            & {\color[HTML]{333333} -}                         & {\color[HTML]{333333} -}          & \cellcolor[HTML]{EFEFEF}{\color[HTML]{333333}75.3}                                                & \cellcolor[HTML]{EFEFEF}{\color[HTML]{333333}64.8}                 \\

{\color[HTML]{333333} SAM BBox 0.75 } & 636 & 0 & -                                                & -    & {\color[HTML]{333333} -}                         & {\color[HTML]{333333} -}                         & \cellcolor[HTML]{EFEFEF}{\color[HTML]{333333} 74.6}                         & \cellcolor[HTML]{EFEFEF}{\color[HTML]{333333} 62.1}            & \cellcolor[HTML]{EFEFEF}{\color[HTML]{333333}27.51}             & {\color[HTML]{333333} -}                         & {\color[HTML]{333333} -}          & \cellcolor[HTML]{EFEFEF}{\color[HTML]{333333}85.7}       
& \cellcolor[HTML]{EFEFEF}{\color[HTML]{333333}74.4}                 \\

{\color[HTML]{333333} MedSAM 1 point} & 636 & 636 & \cellcolor[HTML]{EFEFEF}{\color[HTML]{333333}92.9}                                                & \cellcolor[HTML]{EFEFEF}{\color[HTML]{333333}85.5}    & \cellcolor[HTML]{EFEFEF}{\color[HTML]{333333}82.1}                         & \cellcolor[HTML]{EFEFEF}{\color[HTML]{333333} 73.8}                         & \cellcolor[HTML]{EFEFEF}{\color[HTML]{333333} 81.5}                         & \cellcolor[HTML]{EFEFEF}{\color[HTML]{333333} 74.3}            & \cellcolor[HTML]{EFEFEF}{\color[HTML]{333333}15.68}             & \cellcolor[HTML]{EFEFEF}{\color[HTML]{333333} 81.3}                         & \cellcolor[HTML]{EFEFEF}{\color[HTML]{333333} 74.7}          & \cellcolor[HTML]{EFEFEF}{\color[HTML]{333333}86.8}                                                & \cellcolor[HTML]{EFEFEF}{\color[HTML]{333333}77.5}                 \\

{\color[HTML]{333333} MedSAM 3 points} & 636 & 636 & \cellcolor[HTML]{EFEFEF}{\color[HTML]{333333}93.8}                                                & \cellcolor[HTML]{EFEFEF}{\color[HTML]{333333}86.2}    & \cellcolor[HTML]{EFEFEF}{\color[HTML]{333333} 82.8}                         & \cellcolor[HTML]{EFEFEF}{\color[HTML]{333333} 74.2}                         & \cellcolor[HTML]{EFEFEF}{\color[HTML]{333333} 82.3}                         & \cellcolor[HTML]{EFEFEF}{\color[HTML]{333333} 74.8}            & \cellcolor[HTML]{EFEFEF}{\color[HTML]{333333}15.19}             & \cellcolor[HTML]{EFEFEF}{\color[HTML]{333333} 81.6}                         & \cellcolor[HTML]{EFEFEF}{\color[HTML]{333333} 75.1}          & \cellcolor[HTML]{EFEFEF}{\color[HTML]{333333}87.5 }                                               & \cellcolor[HTML]{EFEFEF}{\color[HTML]{333333}78.6}                 \\ 

{\color[HTML]{333333} MedSAM BBox 0.5} & 636 & 636 & \cellcolor[HTML]{EFEFEF}{\color[HTML]{333333}92.6}                                                & \cellcolor[HTML]{EFEFEF}{\color[HTML]{333333}85.3}    & \cellcolor[HTML]{EFEFEF}{\color[HTML]{333333} 82.0}                         & \cellcolor[HTML]{EFEFEF}{\color[HTML]{333333} 75.2}                         & \cellcolor[HTML]{EFEFEF}{\color[HTML]{333333} 82.0}                         & \cellcolor[HTML]{EFEFEF}{\color[HTML]{333333}74.7}            & \cellcolor[HTML]{EFEFEF}{\color[HTML]{333333}15.05}             & \cellcolor[HTML]{EFEFEF}{\color[HTML]{333333} 82.4}                         & \cellcolor[HTML]{EFEFEF}{\color[HTML]{333333} 75.5}          & \cellcolor[HTML]{EFEFEF}{\color[HTML]{333333}88.5}                                               & \cellcolor[HTML]{EFEFEF}{\color[HTML]{333333}79.2}                 \\

{\color[HTML]{333333} MedSAM BBox 0.75 } & 636 & 636 & \cellcolor[HTML]{EFEFEF}{\color[HTML]{333333}94.6}                                                & \cellcolor[HTML]{EFEFEF}{\color[HTML]{333333}86.7}    & \cellcolor[HTML]{EFEFEF}{\color[HTML]{333333} 82.8}                         & \cellcolor[HTML]{EFEFEF}{\color[HTML]{333333}75.9}                         & \cellcolor[HTML]{EFEFEF}{\color[HTML]{333333}83.6}                         & \cellcolor[HTML]{EFEFEF}{\color[HTML]{333333} 75.6}            & \cellcolor[HTML]{EFEFEF}{\color[HTML]{333333}14.90}             & \cellcolor[HTML]{EFEFEF}{\color[HTML]{333333} 82.8}                         & \cellcolor[HTML]{EFEFEF}{\color[HTML]{333333} 75.7}          & \cellcolor[HTML]{EFEFEF}{\color[HTML]{333333}88.9}                                               & \cellcolor[HTML]{EFEFEF}{\color[HTML]{333333}79.8}                 \\ \hline

Med-SA    1 point       & 636 & 13      & \cellcolor[HTML]{EFEFEF}{\color[HTML]{333333}97.4}                                                & \cellcolor[HTML]{EFEFEF}{\color[HTML]{333333}89.5}    & \cellcolor[HTML]{EFEFEF}{\color[HTML]{333333}86.8}                                       & \cellcolor[HTML]{EFEFEF}{\color[HTML]{333333}78.8}                                       & \cellcolor[HTML]{EFEFEF}{\color[HTML]{333333}89.1}                                       & \cellcolor[HTML]{EFEFEF}{\color[HTML]{333333}81.8}       & \cellcolor[HTML]{EFEFEF}{\color[HTML]{333333}10.38}                                 & \cellcolor[HTML]{EFEFEF}{\color[HTML]{333333}86.3}                                      & \cellcolor[HTML]{EFEFEF}{\color[HTML]{333333}78.7}                           & \cellcolor[HTML]{EFEFEF}{\color[HTML]{333333}92.6}                                                & \cellcolor[HTML]{EFEFEF}{\color[HTML]{333333}84.1}              \\ 

Med-SA    3 points        & 636 & 13      & \cellcolor[HTML]{EFEFEF}{\color[HTML]{333333}97.9}                                                & \cellcolor[HTML]{EFEFEF}{\color[HTML]{333333}89.8}    & \cellcolor[HTML]{EFEFEF}{\color[HTML]{333333}87.1}                                      & \cellcolor[HTML]{EFEFEF}{\color[HTML]{333333}79.0}                                       & \cellcolor[HTML]{EFEFEF}{\color[HTML]{333333}89.8}                                       & \cellcolor[HTML]{EFEFEF}{\color[HTML]{333333}82.3}       & \cellcolor[HTML]{EFEFEF}{\color[HTML]{333333}10.11}                                 & \cellcolor[HTML]{EFEFEF}{\color[HTML]{333333} 86.7}                                       & \cellcolor[HTML]{EFEFEF}{\color[HTML]{333333}79.4}                           & \cellcolor[HTML]{EFEFEF}{\color[HTML]{333333}\textbf{93.4}}                                                & \cellcolor[HTML]{EFEFEF}{\color[HTML]{333333}\textbf{84.7}}              \\ 

Med-SA  BBox 0.5       & 636 & 13      & \cellcolor[HTML]{EFEFEF}{\color[HTML]{333333}97.6}                                                & \cellcolor[HTML]{EFEFEF}{\color[HTML]{333333}89.6}    & \cellcolor[HTML]{EFEFEF}{\color[HTML]{333333}86.4}                                       & \cellcolor[HTML]{EFEFEF}{\color[HTML]{333333}78.5}                                       & \cellcolor[HTML]{EFEFEF}{\color[HTML]{333333}89.5}                                       & \cellcolor[HTML]{EFEFEF}{\color[HTML]{333333}81.9}       & \cellcolor[HTML]{EFEFEF}{\color[HTML]{333333}10.35}                                 & \cellcolor[HTML]{EFEFEF}{\color[HTML]{333333}86.6}                                       & \cellcolor[HTML]{EFEFEF}{\color[HTML]{333333}78.9}                           & \cellcolor[HTML]{EFEFEF}{\color[HTML]{333333}92.1}                                                & \cellcolor[HTML]{EFEFEF}{\color[HTML]{333333}83.0}              \\ 

Med-SA BBox 0.75        & 636 & 13      & \cellcolor[HTML]{EFEFEF}{\color[HTML]{333333}\textbf{98.3}}                                                & \cellcolor[HTML]{EFEFEF}{\color[HTML]{333333}\textbf{90.1}}    & \cellcolor[HTML]{EFEFEF}{\color[HTML]{333333}\textbf{87.5} }                                      & \cellcolor[HTML]{EFEFEF}{\color[HTML]{333333}\textbf{79.9}}                                       & \cellcolor[HTML]{EFEFEF}{\color[HTML]{333333}\textbf{90.5}}                                       & \cellcolor[HTML]{EFEFEF}{\color[HTML]{333333}\textbf{83.0}}       & \cellcolor[HTML]{EFEFEF}{\color[HTML]{333333}\textbf{9.50}}                                 & \cellcolor[HTML]{EFEFEF}{\color[HTML]{333333}\textbf{88.4}}                                       & \cellcolor[HTML]{EFEFEF}{\color[HTML]{333333}\textbf{80.4}}                           & \cellcolor[HTML]{EFEFEF}{\color[HTML]{333333}93.0}                                                & \cellcolor[HTML]{EFEFEF}{\color[HTML]{333333}84.2}              \\ \hline
\end{tabular}}\label{tab:main}
\end{table*}

\begin{figure*}[h]
    \centering
    \includegraphics[width=\linewidth]{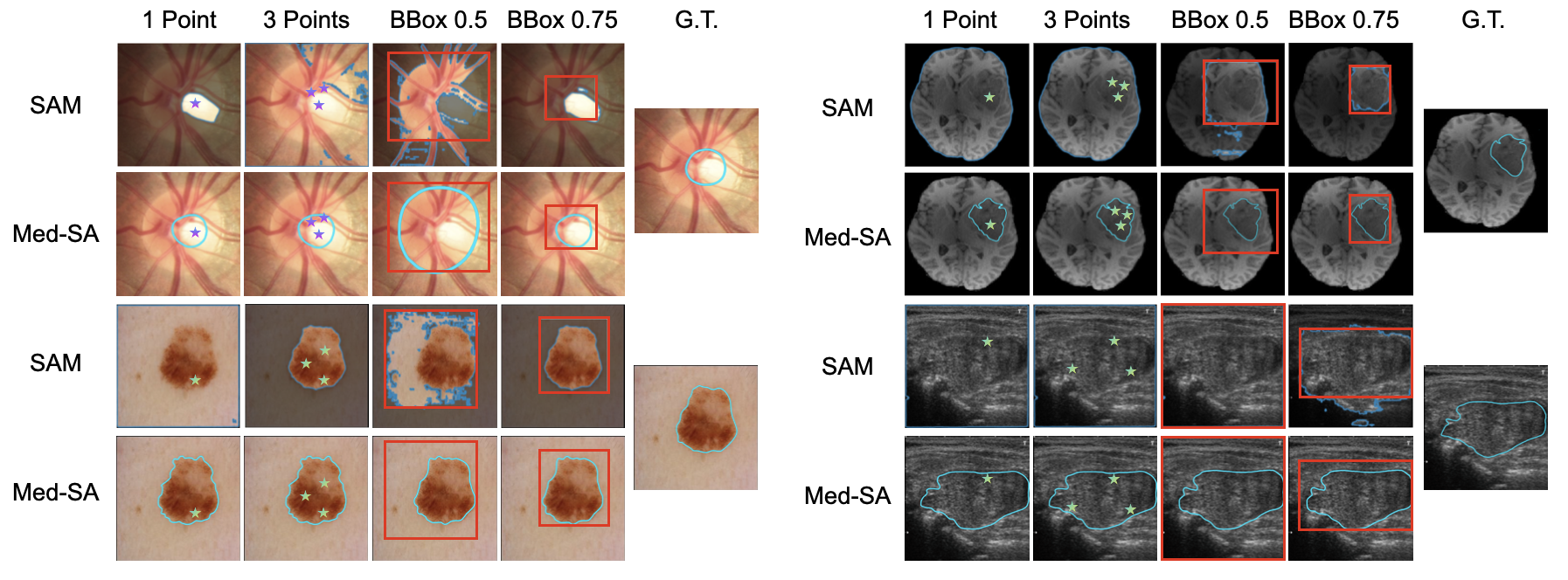}
    \caption{Visual comparison of Med-SA and SAM on medical image segmentation with four different modalities. Top-left: optic disc and cup segmentation from the fundus image. Top-right: brain tumor segmentation from the Brain MRI. Bottom-left: melanoma segmentation from the dermoscopic image. Bottom-right: thyroid nodule segmentation from the ultrasound image.}
    \label{fig:multi}
\end{figure*}

For interactive segmentation, we employ click prompts and bounding box (BBox) prompts during the model training process. To generate BBox prompts, we adopt the same approach as SAM. However, since the original SAM paper provides limited details on click prompt generation, we have devised our own method, which we present here.

The fundamental concept behind our click prompt generation process involves using positive clicks to indicate foreground regions and negative clicks to indicate background regions. We combine random and iterative click sampling strategies to train the model with these prompts. Initially, we utilize random sampling for prompt initialization, and subsequently, we incorporate a few clicks using an iterative sampling procedure. This iterative sampling strategy emulates the interaction with a real user, as each new click is placed in the erroneous region of a prediction generated by the network using the set of previous clicks. We refer to \cite{lin2020interactive} for random sampling generation and \cite{mahadevan2018iteratively} for simulating the iterative sampling process. The detailed implementation can be found in our released code.


\section{Experiments}\label{sec:exp}
\subsection{Dataset}
We conducted experiments on five distinct medical image segmentation datasets, which can be categorized into two types. The first type focused on evaluating general segmentation performance. For this purpose, we selected abdominal multi-organ segmentation, as it represents one of the most significant challenges in medical image segmentation. We utilized the BTCV dataset \cite{fang2020multi}, a widely-used and publicly available benchmark with twelve anatomies as the benchmark.

The other four tasks were used to verify the model's generalization to different modalities, including optic disc and optic cup segmentation over fundus images, brain tumor segmentation over brain MRI images, thyroid nodule segmentation over ultrasound images, and melanoma or nevus segmentation from dermoscopic images. For the fundus image segmentation, we conducted experiments on REFUGE2\cite{fang2022REFUGE2} dataset. For brain tumor segmentation, we conducted experiments on the BraTs 2021 dataset\cite{baid2021rsna}. For thyroid nodule segmentation, we used the TNMIX benchmark, a mixed dataset containing 4554 images from TNSCUI \cite{tnscnn} and 637 images from DDTI \cite{ddti}. Finally, for melanoma or nevus segmentation, we conducted experiments on the ISIC 2019 dataset\cite{milton2019automated}. All datasets are publicly available.

\subsection{Implementation Details}

In this study, we implemented the Med-SA pipeline primarily following the official ViT-H SAM GitHub repository. For 2D medical image training, we adhered to the default training settings of SAM. For 3D medical image training, we used a smaller batch size of 16. For the REFUGE2, TNMIX, and ISIC datasets, we trained the model for 40 epochs. For the BTCV and BraTs datasets, we extended the training to 60 epochs. We chose smaller epoch numbers compared to fully fine-tuned training since we observed that the model converged faster in our setting. In the interactive model, we experimented with four different prompt settings. These included: (1) a random 1 positive point, denoted as "1-point", (2) three positive points, denoted as "3-points", (3) bounding boxes with 50\% overlapping of the target, denoted as "BBox 0.5", and (4) bounding boxes with 75\% overlapping of the target, denoted as "BBox 0.75". All the experiments are implemented with the PyTorch platform and trained/tested on 4 NVIDIA A100 GPUs. We utilized the default settings to reproduce the comparison methods.

\begin{table*}[h]
\centering
\caption{An ablation study on SD-Trans and HyP-Adpt.}
\resizebox{0.95\linewidth}{!}{%
\begin{tabular}{c|c|c|c|cccccc}
\hline
 2D-3D & \multicolumn{3}{c|}{Prompt-Condition} & BTCV & OpticDisc & OpticCup & BrainTumor & ThyroidNodule & Melanoma \\ \hline
\textbf{SD-Trans} & Add & Concat  & \textbf{HyP-Adpt} & Ave-Dice (\%) & Dice (\%)  & Dice (\%)& Dice (\%)&Dice (\%) &Dice (\%) \\ \hline
 & & &              & 79.3 & 90.1     & 80.1     & 77.5       & 76.5  & 89.2        \\ 
\checkmark &  & &  & 84.7  & -  & -     & 81.7       & - & -   \\ 
\checkmark & \checkmark    &     &      &  86.1  & 94.6   & 83.4     & 83.9       & 83.7     & 93.8     \\
\checkmark &   &  \checkmark  &       & 86.4  & 95.7    & 84.0     & 85.1       & 84.8  & 94.5        \\
\checkmark &   &   &  \checkmark      & \textbf{88.3} & \textbf{97.4}   & \textbf{86.8} & \textbf{87.6} & \textbf{86.3} & \textbf{96.3}  \\ \hline     
\end{tabular}}\label{tab:ab}
\end{table*}

\subsection{Comparing with SOTA on Abdominal Multi-organ Segmentation}
To verify the general performance of our proposed Med-SA model, we compare it with SOTA segmentation methods on the multi-organ segmentation datasets BTCV. The quantitative results are presented in \ref{tab:btcv}. In the table, we compare Med-SA with well-recognized medical image segmentation methods, including nnUNet \cite{isensee2021nnu}, TransUNet \cite{chen2021transunet}, UNetr \cite{hatamizadeh2022unetr}, Swin-UNetr \cite{hatamizadeh2022swin}, EnsDiff \cite{wolleb2021diffusion}, and SegDiff \cite{amit2021segdiff}, as well as vanilla SAM and fully fine-turned SAM (MedSAM) \cite{MedSAM}.
We evaluate the segmentation performance using the Dice score.

In the table, we can see that Med-SA achieves a significant improvement over SAM when utilizing only 1-point prompt. Remarkably, on the BTCV dataset, the one-point Med-SA achieves SOTA performance for all 12 organs, surpassing other methods in overall performance. As we provide more fine-grained prompts, the results continue to improve, reaching a final Dice of 89.8\% with BBox 0.75. This result outperforms the previous SOTA (Swin-UNetr) by a significant margin of 2.9\%. Notably, Swin-UNetr consists of 138M turnable parameters, whereas we only update 13M parameters. Surprisingly, we even outperform the fully fine-tuned MedSAM model across all prompt variations. With the proposed SD-Trans and HyP-Adpt, we outperforms MedSAM by updating only 2\% of its total turnable parameters (13M v.s. 636M), which highlights the effectiveness of the proposed techniques. 

When comparing the performance of different prompts in interactive segmentation models (SAM, MedSAM, Med-SA), we notice that 3-points prompts slightly outperform 1-point prompts. 
BBox 0.75 often performs comparably or better than 3-point prompts. However, it is important to note that BBox 0.5 yields subpar performance, indicating the significance of accurate bounding box annotations for achieving performance improvements. All interactive models, including SAM, MedSAM, and Med-SA, exhibit similar behavior across different prompts, demonstrating consistency in their response to prompts.

Considering SAM's performance in \ref{tab:btcv}, we observe that SAM's zero-shot performance is generally inferior to that of fully-trained models (e.g., MedSAM \cite{MedSAM}) in the target medical image segmentation tasks, regardless of the prompt used. While this comparison may seem unfair, as we are comparing SAM's zero-shot performance with fully-trained medical image models, SAM has demonstrated superior zero-shot performance in nature image datasets. This indicates that SAM's zero-shot transferability is less effective for medical images compared to nature image segmentation, which has also been observed in previous studies \cite{deng2023segment,roy2023sam,he2023accuracy}. This finding emphasizes the need for specific techniques to adapt SAM to medical image segmentation.

\ref{fig:amos-vis} presents a qualitative comparison of the performance between Med-SA and SAM. From the figure, it can be observed that Med-SA segments accurately on parts that are difficult to recognize by the human eye. Conversely, SAM fails in many cases where the organ boundaries are visually clear. This further underscores the necessity of fine-tuning a general segmentation model on medical images to achieve optimal performance.

\subsection{Comparing with SOTA on Multi-modality Images}
We also compared Med-SA to specifically optimized segmentation methods across three medical image segmentation tasks with different image modalities. The results are presented in \ref{tab:main}. In the table, ResUnet\cite{yu2019robust} and BEAL\cite{wang2019boundary} are proposed for optic cup segmentation, TransBTS\cite{wang2021transbts} and EnsemDiff\cite{wolleb2021diffusion} are proposed for brain tumor segmentation, MTSeg\cite{gong2021multi} and UltraUNet\cite{chu2021ultrasonic} are proposed for thyroid nodule segmentation, and FAT-Net\cite{wu2022fat} and BAT\cite{wang2021boundary} are proposed for melanoma segmentation. SegDiff, nnUNet, TransUNet, UNetr, and Swin-UNetr are proposed for general medical image segmentation. The segmentation performance was evaluated using Dice score, IoU, and HD95 metrics.

From the table we can see that these specifically optimized methods often perform well within their respective domains but experience drops in performance when applied to other domains. For example, UltraUNet achieves the previous SOTA for thyroid nodule segmentation but performs the worst in optic disc segmentation compared to the other methods. On the other hand, general methods often achieve good results across most modalities but fail to outperform specialized methods in specific tasks such as brain tumor segmentation and thyroid nodule segmentation.

Turning our attention to the interactive models, SAM and MedSAM, we observe that zero-shot SAM struggles with organs/tissues that have ambiguous boundaries in medical images, such as optic disc/cup segmentation or thyroid nodule segmentation. In terms of fully fine-tuned MedSAM, it falls short in brain tumor segmentation due to its limitations in 3D image processing. However, our Med-SA achieves SOTA performance across all segmentation tasks, demonstrating its ability to generalize to various medical segmentation tasks and image modalities. On the widely-used BraTs benchmark, thanks to its adaptability to 3D images, Med-SA outperforms the previous SOTA Swin-UNetr by 2.1\% in Dice score and 1.86 in HD95 metric while utilizing less than 10\% of its turnable parameters.
%


\subsection{Ablation Study}
We conducted a comprehensive ablation study to validate the effectiveness of the proposed SD-Trans and HyP-Adpt. The results are presented in \ref{tab:ab}, where the baseline (first line) represents a simple combination of SAM and the original Adaption method. In the baseline setting, 3D images are treated as a sequence of 2D images and processed individually, without involving prompts in the Adaption process.
As shown in the table, our 2D to 3D design significantly enhances the performance compared to the vanilla SAM plus Adaption setting on both 3D data benchmarks (BTCV and BrainTumor). This improvement highlights the effectiveness of our proposed 2D to 3D design. In the Prompt-conditional Adaption, we compared HyP-Adpt with two simpler alternatives: addition and concatenation, for combining the prompt embedding. While addition and concatenation also show some effectiveness, the improvements achieved are still marginal. On the other hand, using the proposed HyP-Adpt leads to a significant enhancement in performance, further validating the effectiveness of our proposed HyP-Adpt design.


\section{Conclusion}
In this paper, we have extended SAM, a powerful general segmentation model, to address medical image segmentation, introducing Med-SA. Leveraging parameter-efficient adaptation with simple yet effective SD-Trans and HyP-Adpt, we have achieved substantial improvements over the original SAM model. Our approach has resulted in SOTA performance across 17 medical image segmentation tasks spanning 5 different image modalities. We anticipate that this work will serve as a stepping stone towards advancing foundation medical image segmentation and inspire the development of novel fine-tuning techniques.
\clearpage

\fontsize{9.5pt}{10.5pt}
\selectfont
 
\bibliography{aaai22}

\end{document}